\documentclass[10pt,twocolumn,letterpaper]{article}

\usepackage[pagenumbers]{cvpr} 

\usepackage{graphicx}
\usepackage{amsmath}
\usepackage{amssymb}
\usepackage{booktabs}
\usepackage{tabularx}
\usepackage{colortbl}
\usepackage[table]{xcolor}

\usepackage[pagebackref,breaklinks,colorlinks]{hyperref}

\usepackage[capitalize]{cleveref}
\crefname{section}{Sec.}{Secs.}
\Crefname{section}{Section}{Sections}
\Crefname{table}{Table}{Tables}
\crefname{table}{Tab.}{Tabs.}

\def\cvprPaperID{*****} 
\def\confName{CVPR}
\def\confYear{2023}

\begin{document}

\title{\LaTeX\ Author Guidelines for \confName~Proceedings}

\title{Tracking Anything in High Quality}

\author{Jiawen Zhu$^1$, \ Zhenyu Chen$^1$\protect, Zeqi Hao$^1$, Shijie Chang$^1$, Lu Zhang$^1$, Dong Wang$^1$, Huchuan Lu$^{1}$, \\
Bin Luo$^{2}$,  Jun-Yan He$^{2}$, Jin-Peng Lan$^{2}$, Hanyuan Chen$^{2}$, Chenyang Li$^{2}$\\
$^1$Dalian University of Technology, China \quad
$^2$DAMO Academy, Alibaba Group \\
{\tt\small \{jiawen,dlutczy,hzq,csj\}@mail.dlut.edu.cn, \{luzhangdut,junyanhe1989,lanjinpeng1015\}@gmail.com} \\
{\tt\small \{wdice,lhchuan\}@dlut.edu.cn, \{luwu.lb,hanyuan.chy,lee.lcy\}@alibaba-inc.com}
}

\maketitle

\begin{abstract}
	Visual object tracking is a fundamental video task in computer vision.
	Recently, the notably increasing power of perception algorithms allows the unification of single/multi-object and box/mask-based tracking.
	Among them, the Segment Anything Model (SAM) attracts much attention.
	In this report, we propose HQTrack, a framework for \textbf{H}igh \textbf{Q}uality \textbf{Track}ing anything in videos. HQTrack mainly consists of a video multi-object segmenter (VMOS) and a mask refiner (MR).
	Given the object to be tracked in the initial frame of a video, VMOS propagates the object masks to the current frame.
	The mask results at this stage are not accurate enough since VMOS is trained on several close-set video object segmentation (VOS) datasets, which has limited ability to generalize to complex and corner scenes.
	To further improve the quality of tracking masks, a pre-trained MR model is employed to refine the tracking results.
	As a compelling testament to the effectiveness of our paradigm, 
	without employing any tricks such as test-time data augmentations and model ensemble,
	HQTrack ranks the 2$^{nd}$ place in the Visual Object Tracking and Segmentation (VOTS2023) challenge.
	Code and models are available at \href{https://github.com/jiawen-zhu/HQTrack}{https://github.com/jiawen-zhu/HQTrack}.
 \end{abstract}

\section{Introduction}
\label{sec:intro}
As a fundamental video task in computer vision, visual object tracking has become the cornerstone of many related areas, such as robot vision and autonomous driving. The task aims at consistently locating the specified object in a video sequence. As one of the most influential challenges in the tracking field, Visual Object Tracking (VOT) challenge \cite{vot21,vot22} attracted a lot of attention, and many SOTA algorithms participate to show their cutting-edge performance. The Visual Object Tracking and Segmentation challenge (VOTS2023\footnote{\url{https://www.votchallenge.net/vots2023}}) relaxes the constraints enforced in past VOT challenges for considering general object tracking in a broader context. Therefore, VOTS2023 merges short-term and long-term, single-target and multiple-target tracking with segmentation as the only target location specification. This poses more challenges, \eg, inter-object relationship understanding, multi-target trajectory tracking, accurate mask estimation, etc.

Visual object tracking has made great strides with deep learning techniques\cite{alexnet, resnet, vit}. 
Previous methods can be grouped into either online-update trackers\cite{eco, dimp} and Siamese trackers\cite{siamesefc, siamrcnn}. 
Recently, Transformer\cite{attention_is_all} sweeps in computer vision, the dominant tracking methods are Transformer-based trackers\cite{transt,stark,mixformer,ostrack}. TransT\cite{transt} proposes transformer-based ECA and CFA modules to replace the long-used correlation calculation.
Benefiting from Transformer's superior long-range modeling capability, TransT outperforms the previous correlation modules which are a capable class of linear modeling. 
More recently, some trackers\cite{mixformer, ostrack} introduce pure transformer architecture, and the feature extracting and template-search region interaction is completed in a single backbone, tracking performance is pushed to new records.
These trackers mainly focus on single object tracking and output the bounding box for performance evaluation.
Hence, merely employing SOT trackers is not well-suited to the VOTS2023 challenge.

Video object segmentation aims to segment out the specific objects of interest in a video sequence. Similar to VOT, semi-supervised video object segmentation also manually provides the annotated in the first frame. The main difference is that the VOS task provides a more fine-grained mask annotation. 
Early VOS methods propagate object masks over video frames via motion clues\cite{segflow,tsai2016video} or adopt online learning strategies\cite{caelles2017one,li2018video}.
Recently, Space-Temporal Memory (STM) network~\cite{oh2019video, wang2017learning} extracts the spatio-temporal context from a memory bank to handle the appearance changes and occlusions, offering a promising solution for semi-supervised video object segmentation.
For multi-object segmentation, these methods segment the objects one by one, the final results are merged masks by post ensemble.
AOT\cite{aot} proposes an identification mechanism that can encode, match, and segment multiple objects at the same time. Based on AOT\cite{aot}, DeAOT\cite{deaot} decouples the hierarchical propagation of object-agnostic and object-specific embeddings from previous frames to the current frame, further improving the VOS accuracy.

Although the above VOS methods can handle tracking task with multi-object and mask output, 
challenges in VOTS2023 benchmark remain. 
\textbf{\textit{(i)}} VOTS videos contain a large number of long-term sequences, the longest of which exceeds 10,000 frames, which requires the tracker to be able to discriminate the drastic changes in object appearance and adapt to variations in the environment.
At the same time, long-term video sequences also make some memory-based methods face memory bank space challenges.
\textbf{\textit{(ii)}} In VOTS videos, targets will leave the field of view and then returns.
Trackers require additional design to accommodate the disappearance and appearance of targets. 
\textbf{\textit{(iii)}} A series of challenges such as fast motion, frequent occlusion, distractors, and tiny objects also make this task more difficult. 

In this work, we propose Tracking Anything in High Quality (termed HQTrack), which mainly consists of a video multi-object segmenter (VMOS) and a mask refiner (MR). VMOS is an improved variant of DeAOT~\cite{deaot}, we cascade a 1/8 scale gated propagation module (GPM) for perceiving small objects in complex scenarios. Besides, Intern-T\cite{internimage} is employed as our feature extractor to enhance object discrimination capabilities.
To save memory usage, a fixed length of long-term memory is used in VMOS, excluding the initial frame, the memory of early frames will be discarded.
On the other hand, it should be beneficial to apply a large segmentation model to refine our tracking masks.
SAM\cite{sam} is prone to failure when predicting objects with complex structures\cite{sam_hq}, and these difficult cases are common in VOTS chanllenge.
To further improve the quality of tracking masks, a pre-trained HQ-SAM\cite{sam_hq} model is employed to refine  the tracking masks. 
We calculate the outer enclosing boxes of the predicted masks from VMOS as box prompts and feed them into HQ-SAM together with the original images to gain the refined masks, the final tracking results are selected from VMOS and MR.

Finally, HQTrack obtains an impressive 0.615 quality score on the VOTS2023 test set, achieving runner-up at the VOTS2023 challenge.

\begin{figure*}[ht]
    \vspace{-2mm}
        \centering
        \includegraphics[width=0.85\linewidth]{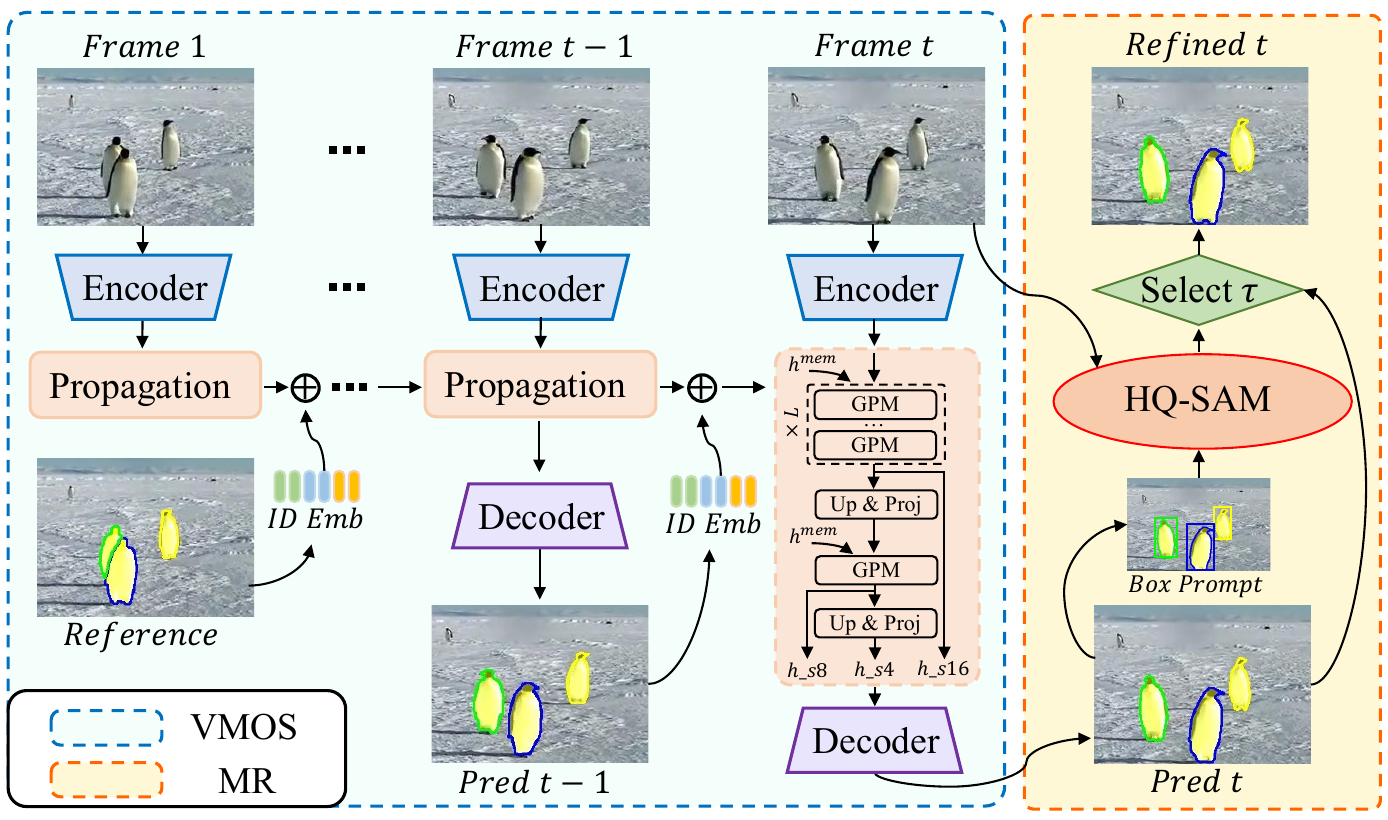}
        \vspace{-1.5mm}
        \caption{Overview of HQTrack. It mainly consists of a video multi-object segmenter (VMOS) and a mask refiner (MR).}
        \label{fig:method_overview}
    \vspace{-2mm}
\end{figure*}

\section{Method}
In this section,  we present our HQTrack in detail. 
We first showcase the pipeline of our method. Subsequently, we introduce each component in our framework. Finally, we describe the training and inference details. 

\subsection{Pipeline}
\label{sec:pipline}

The pipeline of the proposed HQTrack is depicted in Figure~\ref{fig:method_overview}.
Given a video and the first frame reference (mask annotated), 
HQTrack first segments the target objects for each frame via VMOS.
The segmentation results of the current frame are from the propagation of the first frame along the temporal dimension, utilizing the modeling of appearance/identification information and long/short-term memory.
VMOS is a variant of DeAOT\cite{deaot} so that it can accomplish the modeling of multiple objects in a scene within a single propagation process.
Furthermore, we employ HQ-SAM\cite{sam_hq} as our MR to refine the segmentation masks of VMOS.
HQ-SAM is a variant of SAM\cite{sam}, it can handle objects with more complex structures than SAM.
We first perform bounding box extraction on the target masks predicted by VMOS, and they are fed into the HQ-SAM model as box prompts.
Last, we design a mask selector to select the final results from VMOS and MR.

\subsection{Video Multi-object Segmenter (VMOS))}
\label{sec:vmos}
VMOS is a variant of DeAOT\cite{deaot}, thereby in this subsection, we first provide a brief revisiting of DeAOT which is the baseline of our VMOS, then we delve into the design of our VMOS.

\noindent\textbf{DeAOT.}
AOT\cite{aot} proposes to incorporate an identification mechanism to associate multiple objects in a unified embedding space which enables it to handle multiple objects in a single propagation. 
DeAOT is a video object segmentation model with a AOT-like hierarchical propagation. To alleviate the loss of object-agnostic visual information in deep propagation layers, DeAOT proposes to decouple the propagation of visual and identification embeddings into a dual-branch gated propagation module (GPM).
GPM is an efficient module with single-head attention for constructing hierarchical propagation.

\noindent\textbf{VMOS.}
The video multiple object segmenter (VMOS) in HQTrack is a variant of DeAOT.
As shown in the left of Figure~\ref{fig:method_overview}, to improve the segmentation performance, especially perceiving tiny objects,
we cascade a GPM with 8$\times$ scale and expand the propagation process to multiple scales.
The original DeAOT only performs propagating operation on the visual and identification features of 16$\times$ scale. 
At this scale, lots of detailed object clues are lost, especially for tiny objects, 16$\times$ scale features are insufficient for accurate video object segmentation.
In our VMOS, considering the memory usage and model efficiency, we only use up-sampling and linear projection to upscale the propagation features to 4$\times$ scale.
Multi-scale propagation features will be fed into the decoder along with multi-scale encoder features for mask prediction.
Decoder is a simple FPN~\cite{fpn}.
In addition, as a new large-scale CNN-based foundation model, Internimage~\cite{internimage} 
employs deformable convolution as the core operator, showing impressive performance on various representative tasks \eg, object detection and segmentation.
In VMOS, Intern-T is employed as our encoder to enhance object discrimination capabilities. 

\subsection{Mask Refiner (MR)}
\label{sec:mr}

MR is a pre-trained HQ-SAM~\cite{sam_hq}, in this section, we first revisit the HQ-SAM method which is a variant of SAM~\cite{sam}, then we provide the usage of HQ-SAM.

\noindent\textbf{SAM and HQ-SAM.}
Segment anything model (SAM) has recently attracted high-heat attention in the field of image segmentation, and researchers have utilized SAM to secondary a series of work (including but not limited to segmentation) with many stunning results.
SAM scales up segmentation models by training with a high-quality annotated dataset containing 1.1 billion masks.
In addition to the powerful zero-shot capabilities brought  by large-scale training, SAM also involves flexible human interaction mechanisms achieved by different prompt formats.  
However, when the processed image contains  objects with intricate structures, SAM's prediction masks tend to fall short.
To tackle such an issue as well as maintain  SAM's original promptable design, efficiency, and zero-shot generalizability, Ke~\etal propose HQ-SAM~\cite{sam_hq}.  HQ-SAM introduces a few additional parameters to the pre-trained SAM model. High-quality mask is obtained by injecting a learning output token into SAM's mask decoder.

\noindent\textbf{MR.} HQTrack employs the above HQ-SAM as our mask refiner.
As shown in the right of Figure~\ref{fig:method_overview}, we take the prediction mask from VMOS as the input of MR.
Since the VMOS model is trained on scale-limited close-set datasets, the first stage mask from VMOS probably with insufficient quality especially handling some complex scenarios.
Hence, employing a large-scale trained segmentation algorithm to refine the primary segmentation results will bring considerable performance improvement.
Specifically, we calculate the outer enclosing boxes of the predicted mask from VMOS as the box prompts and feed them into HQ-SAM together with the original image to obtain the refined masks. HQ-SAM here is a version with a ViT-H backbone.
Finally, the output mask of HQTrack is selected from the mask results from VMOS and HQ-SAM. 
Specifically, we find that for the same target object, the mask refined by HQ-SAM is sometimes completely different from the predicted mask of VMOS (very low IoU score) which instead harms the segmentation performance. This may be a result of the different understanding and definition of object between HQ-SAM and reference annotation. Hence, we set an IoU threshold $\tau$ (between masks from VMOS and HQ-SAM) to determine which mask will be used as the final output. In our case, when the IoU score is higher than $\tau$, we choose the refined mask. This process constrains HQ-SAM to focus on refining the current object mask rather than re-predicting another target object.

\

\section{Implementation Details}

In VMOS of HQTrack, InternImage-T~\cite{internimage} is employed as the backbone for the image encoder for the trade-off between accuracy and efficiency.
The layers number of the GMP for 16$\times$ and 8$\times$ scale is set to 3 and 1. The 4$\times$ scale propagation features are up-sampled and projection features from 8$\times$ scale.
The long and short-term memory is used in our segmenter to deal with object appearance changes in long-term video sequences.
To save memory usage, we use a fixed length of long-term memory of 8, excluding the initial frame, the early memory will be discarded. 

\noindent\textbf{Model Training.}
The training process comprises two stages, following previous methods~\cite{aot,deaot}. 
In the first phase, we pre-train VMOS on synthetic video sequences generated from static image datasets~\cite{cheng2014global,everingham2010pascal,hariharan2011semantic,lin2014microsoft,shi2015hierarchical}.  
In the second stage, VMOS uses multi-object segmentation datasets for training for a better understanding of the relationship between multiple objects. The training splits of DAVIS~\cite{davis}, YoutubeVOS~\cite{youtubevos}, VIPSeg~\cite{vipseg}, BURST~\cite{burst}, MOTS~\cite{mots}, and OVIS~\cite{ovis} are chosen for training our VMOS, in which OVIS is employed to improve the robustness of the tracker in handling occluded objects. 
We use 2 NVIDIA Tesla A100 GPUs with a global batch size of 16 to train our VMOS.
The pre-training stage uses an initial learning rate of $4 \times 10^{-4}$
for 100,000 steps.
The second stage uses an initial learning rate of $2 \times 10^{-4}$ for 150,000 steps. Learning rates gradually decay to $1 \times 10^{-5}$ in a polynomial manner~\cite{yang2020collaborative}.

\noindent\textbf{Inference.}
The inference process is as described in our pipeline. 
We do not use any test time augmentation (TTA) such as flipping, multi-scale testing, and model ensemble.

\section{Experiment}
\label{sec:challenge}

\begin{table}[t]
    \renewcommand\arraystretch {1.25}
    \centering
    \small
    \setlength{\tabcolsep}{1pt} %
    \begin{tabularx}{\linewidth}{>{\raggedright\arraybackslash}p{2.8cm} >{\centering\arraybackslash}X>{\centering\arraybackslash}X>{\centering\arraybackslash}X>{\centering\arraybackslash}X>{\centering\arraybackslash}X>{\centering\arraybackslash}X}
  \hline
  Method & AUC &A &R &NRE$\downarrow$ &DRE$\downarrow$ &ADQ   \\
  \hline
  MS\_AOT (Separate) &0.552&0.625&0.831&0.063&0.106&0.417  \\
  MS\_AOT (Joint) &\textbf{0.566}&0.645&0.782&0.097&0.121&0.561\\
  \hline
    \end{tabularx}
    \vspace{-2mm}
    \caption{Ablation study of separate tracking $v.s.$ joint tracking paradigm on VOTS2023 validation set. 
    The metrics marked with	$\downarrow$ indicate that smaller is better and vice versa. 
	NRE: Not-Reported Error. DRE: Drift-Rate Error. ADQ: Absence-Detection Quality. 
	We refer readers to ~\cite{vots2023_metric} for more details about evaluation metrics.
}
    \label{table:separate_joint}
\end{table}

\begin{table}[t]
	\renewcommand\arraystretch {1.25}
	\centering
	\small
	\setlength{\tabcolsep}{1pt} %
	\begin{tabularx}{\linewidth}{>{\centering\arraybackslash}p{4mm}|>{\raggedright\arraybackslash}p{2.35cm} >{\centering\arraybackslash}X>{\centering\arraybackslash}X>{\centering\arraybackslash}X>{\centering\arraybackslash}X>{\centering\arraybackslash}X>{\centering\arraybackslash}X}
		\hline
		\# &Method & AUC &A &R &NRE$\downarrow$ &DRE$\downarrow$ &ADQ   \\
		\hline
		1& Baseline & 0.576	&0.675	&0.77	&0.122	&0.108	&0.581  \\
		2& $w/$ InternImage-T &0.611	&0.656	&0.809	&0.137	&0.054	&0.788\\
		3& VMOS &\textbf{0.650}	&0.681	&0.886	&0.059	&0.055	&0.648\\
		\hline
	\end{tabularx}
	\vspace{-2mm}
	\caption{Ablation study of components of VMOS on VOTS2023 validation set. We train a DeAOT~\cite{deaot} as the baseline method.}
	\label{table:vmos_com}
\end{table}

\begin{table}[t]
	\renewcommand\arraystretch {1.25}
	\centering
	\small
	\setlength{\tabcolsep}{1pt} %
	\begin{tabularx}{\linewidth}{>{\centering\arraybackslash}p{1cm} >{\centering\arraybackslash}X>{\centering\arraybackslash}X>{\centering\arraybackslash}X>{\centering\arraybackslash}X>{\centering\arraybackslash}X>{\centering\arraybackslash}X}
		\hline  
		\rowcolor{white!}
		\hline 
		$G=$ & AUC &A &R &NRE$\downarrow$ &DRE$\downarrow$ &ADQ   \\
		\hline
		10 &0.610	&0.668	&0.807	&0.110	&0.083	&0.694  \\\rowcolor{gray!25}
		20 &0.607	&0.65	&0.806	&0.12	&0.074	&0.697\\
		30 &0.626	&0.689	&0.813	&0.127	&0.060	&0.715\\\rowcolor{gray!25}
		40 &0.650	&0.681	&0.886	&0.059	&0.055	&0.648\\
		50 &\textbf{0.669}	&0.692	&0.885	&0.057	&0.058	&0.682\\\rowcolor{gray!25}
		60 &0.653	&0.669	&0.889	&0.059	&0.052	&0.685\\
		70 &0.656	&0.688	&0.865	&0.052	&0.082	&0.666\\\rowcolor{gray!25}
		\hline
	\end{tabularx}
	\vspace{-2mm}
	\caption{Ablation study of long-term memory gap ($G$) on VOTS2023 validation set.}
	\label{table:ltm_param}
	\vspace{-5mm}
\end{table}

\subsection{Ablation Study}

\noindent\textbf{Separate tracking $v.s.$ Joint tracking.}
We conduct ablation studies on different tracking paradigms. 
\textit{Separate tracking} means initializing a separate tracker for each target object, and running multiple times of inference for multiple object tracking.
\textit{Joint tracking} means joint tracking all target objects with a single tracker.
We choose MS\_AOT~\cite{vot22} (removing Mixformer~\cite{mixformer}) as the baseline.
The results on VOTS2023 validation set are shown in Tabled~\ref{table:separate_joint}.
We can see that joint tracking shows better performance than separate tracking.
It may be that when joint tracking, the tracker will have a better understanding of the relationship between the target objects which makes the tracker obtain better robustness to distractor interference.

\noindent\textbf{Component-Wise Analysis on VMOS.}
Table~\ref{table:vmos_com} shows the component-wise study results on VMOS.
\#1 is a trained baseline method DeAOT~\cite{deaot}.
In \#2, we replace the original ResNet50~\cite{resnet} backbone with InternImage-T~\cite{internimage}, and the AUC score increases to 0.611.
Then, as reported in \#3, we add the multi-scale propagation mechanism as described in Section~\ref{sec:vmos}, the performance boosts to 0.650 in terms of AUC score, with a remarkable improvement of 3.9\%, which demonstrates the effectiveness.

\noindent\textbf{Long-term Memory Gap.}
Since the VOTS video sequences tend to be long (the longest exceeds 10,000 frames), the original long-term memory gap parameter on test time for the VOS benchmark is less suitable. 
Therefore, we do an ablution study on long-term memory gap ($G$) parameter as shown in Table~\ref{table:ltm_param}. 
We find that a memory gap of 50 shows the best performance.

\begin{figure}[!t]
	\vspace{-2mm}
	\centering
	\includegraphics[width=0.76\linewidth]{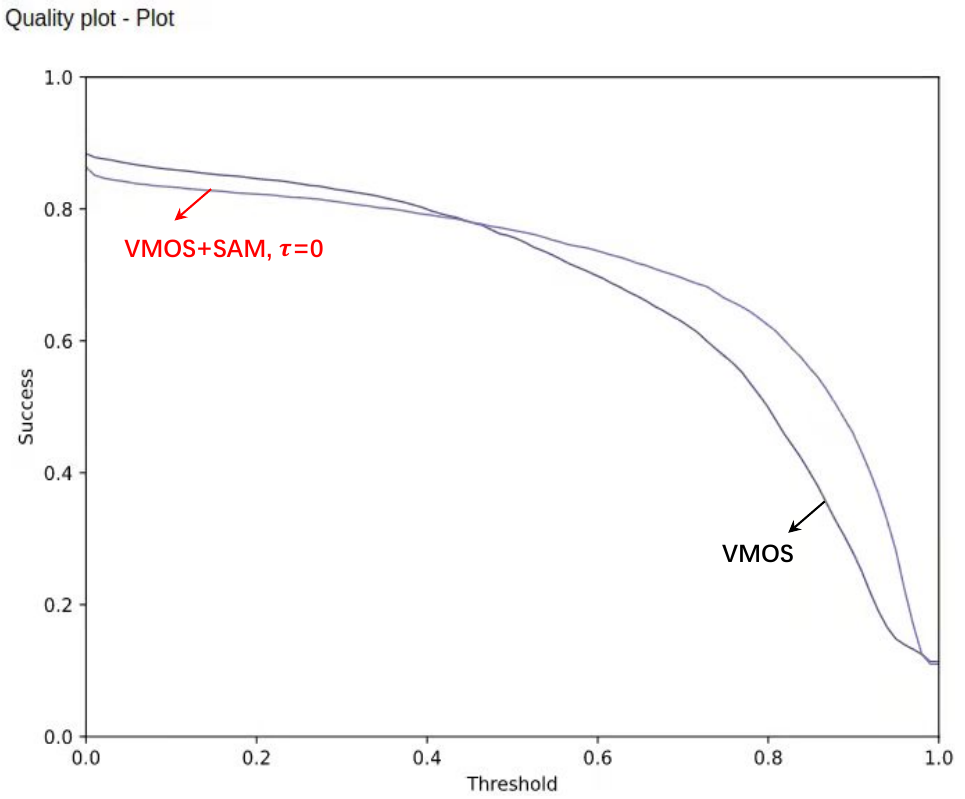}
	\vspace{-1.5mm}
	\caption{VMOS $v.s.$ VMOS + SAM on VOST2023 validation set. SAM is employed to refine all the masks from VMOS.}
	\label{fig:sam_plot}
	\vspace{-2mm}
\end{figure}

\noindent\textbf{Analysis on Mask Refiner (MR).}
As we discuss in Section~\ref{sec:mr}, directly refining all the segmentation masks is not optimal. We provide a comparison between VMOS and VMOS + SAM in Figure~\ref{fig:sam_plot}. In VMOS + SAM case, a SAM-h~\cite{sam} is employed to refine all the object masks from VMOS. 
We can see that refining by SAM can bring significant improvement. However, for these masks with low quality (with low IoU score on ground truth), SAM harms the performance instead.
Therefore, we propose to select mask results from VMOS and SAM. We calculate the IoU score between the masks from VMOS and SAM. When the IoU score is higher than $\tau$, we choose the refined mask as the final output. 
We evaluate the influence of threshold $\tau$ in MR on the VOTS2023 validation set, the results are shown in Table~\ref{table:mr_iou}.
$\tau=0.1$ yields the most promising results and we choose this setting in HQTrack.

\begin{figure*}[ht]
	\vspace{-2mm}
	\centering
	\includegraphics[width=0.95\linewidth]{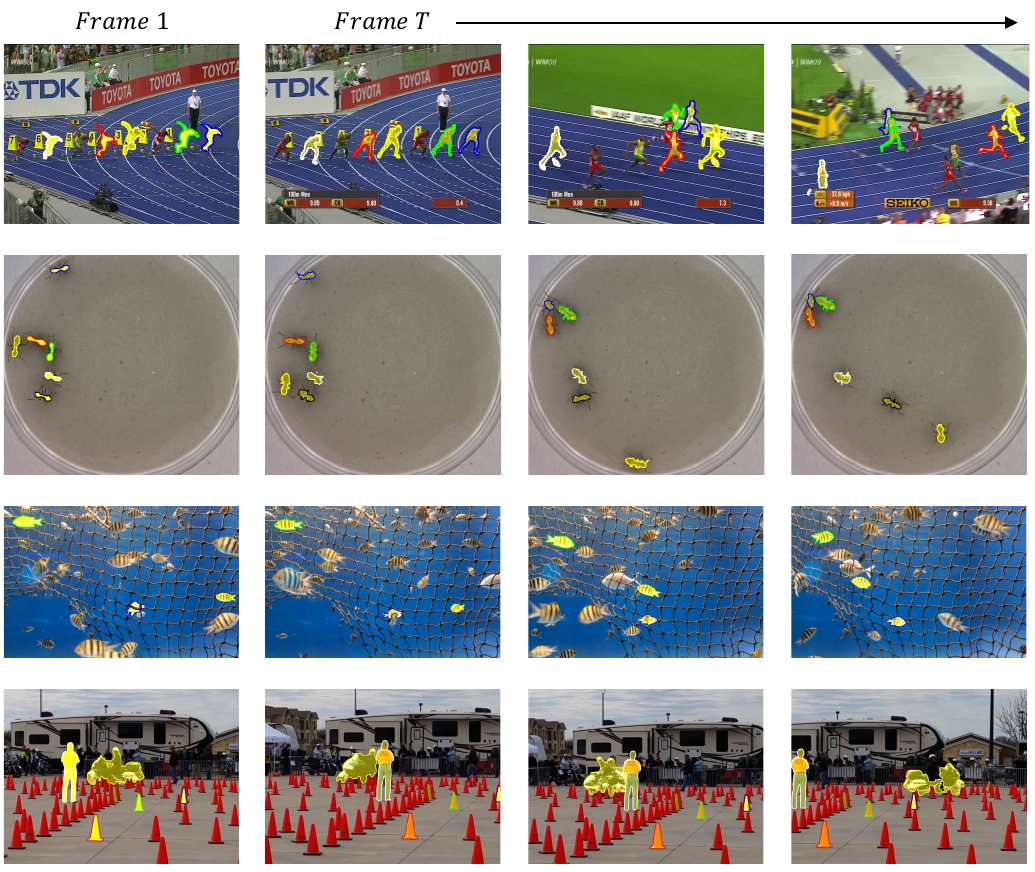}
	\vspace{-1.5mm}
	\caption{Qualitative results of HQTrack on videos from VOTS2023 test set. }
	\label{fig:vis}
	\vspace{-2mm}
\end{figure*}

\begin{table}[t]
	\renewcommand\arraystretch {1.25}
	\centering
	\small
	\setlength{\tabcolsep}{1pt} %
	\begin{tabularx}{\linewidth}{>{\centering\arraybackslash}p{1cm} >{\centering\arraybackslash}X>{\centering\arraybackslash}X>{\centering\arraybackslash}X>{\centering\arraybackslash}X>{\centering\arraybackslash}X>{\centering\arraybackslash}X}
		\hline  
		\rowcolor{white!}
		\hline 
		$\tau=$ & AUC &A &R &NRE$\downarrow$ &DRE$\downarrow$ &ADQ   \\
		\hline
		0 &0.702	&0.756	&0.866	&0.072	&0.062	&0.769  \\\rowcolor{gray!25}
		0.1 &\textbf{0.708}	&0.753	&0.878	&0.072	&0.050	&0.769\\
		0.2 &0.707	&0.753	&0.878	&0.072	&0.050	&0.768\\\rowcolor{gray!25}
		0.3 &0.704	&0.750	&0.878	&0.072	&0.050	&0.764\\
		0.4 &0.701	&0.745	&0.878	&0.072	&0.050	&0.763\\\rowcolor{gray!25}
		0.5 &0.695	&0.739	&0.878	&0.072	&0.050	&0.758\\
		\hline
	\end{tabularx}
	\vspace{-2mm}
	\caption{Tracking performance with different threshold $\tau$ on VOTS2023 validation set. Mask refiner (MR) is a SAM\_H model.}
	\label{table:mr_iou}
\end{table}

\begin{table}[t]
	\renewcommand\arraystretch {1.25}
	\centering
	\small
	\setlength{\tabcolsep}{1pt} %
	\begin{tabularx}{\linewidth}{>{\centering\arraybackslash}p{2.5cm} >{\centering\arraybackslash}X>{\centering\arraybackslash}X>{\centering\arraybackslash}X>{\centering\arraybackslash}X>{\centering\arraybackslash}X>{\centering\arraybackslash}X}
		\hline
		Method & AUC &A &R &NRE$\downarrow$ &DRE$\downarrow$ &ADQ   \\
		\hline
		VMOS (Res50) &0.564	&0.693	&0.759	&0.155	&0.086	&0.691  \\
		VMOS &0.596	&0.724	&0.765	&0.159	&0.075	&0.711  \\
		VMOS + SAM\_H &0.610 &0.751	&0.757  &0.159  &0.084 &0.706\\
		\textbf{HQTrack} &\textbf{0.615} &0.752 &0.766  &0.155 &0.079	&0.694
		\\
		\hline
	\end{tabularx}
	\vspace{-2mm}
	\caption{Performance on VOTS2023 test set.}
	\label{table:testset_results}
	\vspace{-5mm}
\end{table}

\subsection{Challenge Results}
\begin{figure}[!t]
	\vspace{-2mm}
	\centering
	\includegraphics[width=0.8\linewidth]{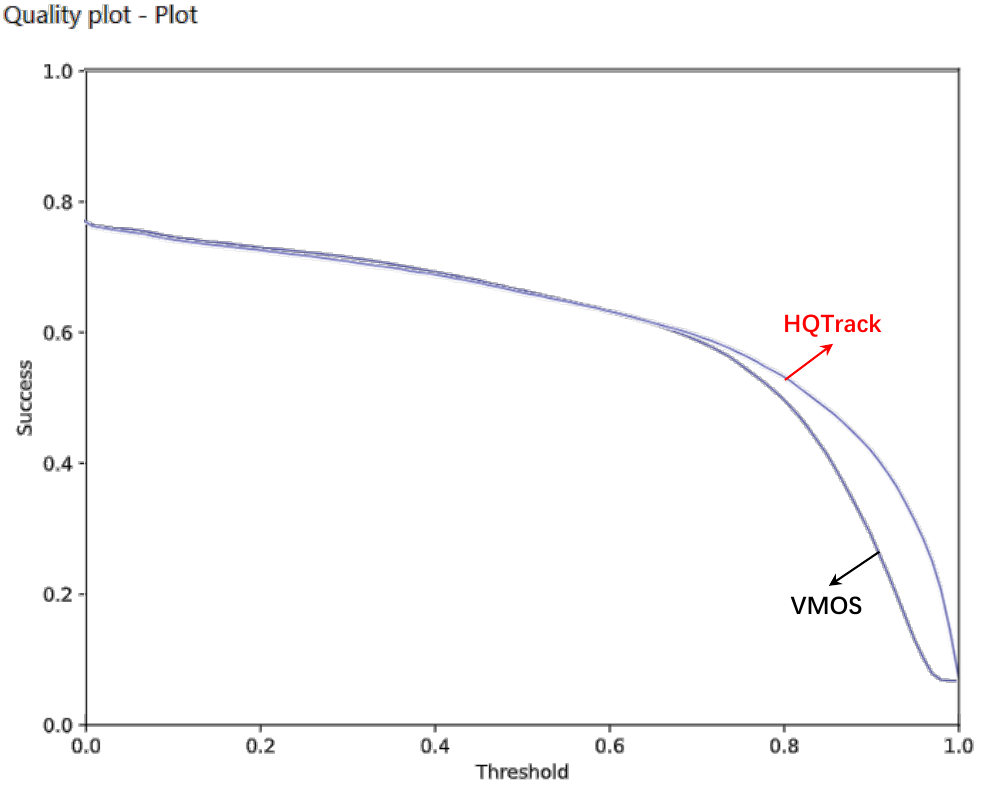}
	\vspace{-1.5mm}
	\caption{VMOS $v.s.$ HQTrack on VOST2023 test set. }
	\label{fig:hqsam_plot}
	\vspace{-2mm}
\end{figure}

The results on VOTS2023 test set are shown in Table~\ref{table:testset_results}.
After replacing the VMOS encoder from ResNet50~\cite{resnet} to InternImage-T~\cite{internimage}, the AUC score increased  by 3.2\%.
When using SAM\_H to refine the masks of VMOS, the performance in terms of AUC increased by 1.4\%. 
After employing HQ-SAM\_H as our mask refine module, the AUC score boosts to 0.615, which outperforms VMOS by 0.9\%.
Figure~\ref{fig:hqsam_plot} provides the quality plot comparison between VMOS and HQtrack.
As we can see and compare with Figure~\ref{fig:sam_plot}, selectively taking the processed results of the MR can effectively avoid performance degradation from low IoU objects. 
Finally, HQTrack ranks 2nd place\footnote{\url{https://eu.aihub.ml/competitions/201\#results}, VOTS2023 benchmark is open for allowing post-challenge submissions.} in the Visual Object Tracking and Segmentation Challenge.

\subsection{Visualization}
Figure~\ref{fig:vis} provides some representative visual results on challenging video sequences. 
As shown, HQTrack demonstrates strong tracking capabilities.
It can stably handle long-term object tracking scenarios, tracking multiple objects at the same time, and capturing target objects accurately even if there are a lot of distractors. 
With the help of HQ-SAM, accurate masks can also be segmented when facing challenges such as object appearance changes, fast motion, and scale changes.

\section{Conclusion}

In this report, we propose Tracking Anything in High Quality (HQTrack).
HQTrack mainly consists of a video multi-object segmenter (VMOS) and a mask refiner (MR).
VMOS is responsible for propagating multiple targets in video frames, and MR is a large-scale pre-trained segmentation model in charge of refining the segmentation masks.
HQTrack demonstrates powerful object tracking and segmentation capabilities.
Finally, HQTrack achieves the 2nd place in the Visual Object Tracking and Segmentation (VOTS2023) challenge.

{\small
\bibliographystyle{ieee_fullname}
\bibliography{main}
}

\end{document}